%% file: src/main.tex
\pdfoutput=1
\documentclass[11pt]{article}

\input{src/preamble}

\begin{document}

\input{src/intro}

\input{src/algorithm}

\input{src/validation}

\input{src/ec}

\input{src/outro}

\nocite{boldt2024review}
\bibliography{src/main}

\input{src/appendix}

\typeout{INFO: \arabic{comment} comments.}
\end{document}

%% file: src/preamble.tex
\usepackage[preprint]{acl-style-files/latex/acl}
\usepackage{times}
\usepackage{latexsym}
\usepackage[T1]{fontenc}
\usepackage[utf8]{inputenc}
\usepackage{microtype}
\usepackage{inconsolata}
\usepackage{graphicx}
\usepackage{amssymb,amsmath}
\usepackage{wasysym}

\title{Morpheme Induction for Emergent Language}

\author{Brendon Boldt \and David Mortensen \\
  Language Technologies Institute \\
  Carnegie Mellon University \\
  Pittsburgh, PA, USA \\
  \texttt{\{bboldt,dmortens\}@cs.cmu.edu}}

\usepackage{hyperref}
\usepackage{xurl}
\usepackage[table]{xcolor}
\usepackage{latexsym,amsmath}
\usepackage{booktabs}
\usepackage{multirow}
\usepackage{enumitem}
\usepackage[nameinlink,capitalize]{cleveref}
\usepackage{pgfplots}
\usepackage{tikz}
\usetikzlibrary{shapes,arrows,shadows.blur,calc,positioning,fit,backgrounds}
\usepackage{algorithm,listings}

\newcounter{comment}

\everymath=\expandafter{\the\everymath\displaystyle}

\ifdefined\pdftexversion\else  
  \usepackage{fontspec}
\fi
\makeatletter\@ifpackageloaded{underscore}{}{\usepackage[strings]{underscore}}\makeatother

%% file: src/intro.tex
\maketitle

\begin{abstract}
We introduce CSAR, an algorithm for inducing morphemes from emergent language corpora of parallel utterances and meanings.
It is a greedy algorithm that (1) weights morphemes based on mutual information between forms and meanings, (2) selects the highest-weighted pair, (3) removes it from the corpus, and (4) repeats the process to induce further morphemes (i.e., Count, Select, Ablate, Repeat).
The effectiveness of CSAR is first validated on procedurally generated datasets and compared against baselines for related tasks.
Second, we validate CSAR's performance on human language data to show that the algorithm makes reasonable predictions in adjacent domains.
Finally, we analyze a handful of emergent languages, quantifying linguistic characteristics like degree of synonymy and polysemy.
\end{abstract}

\section{Introduction}

Emergent languages---communication systems invented by neural networks via reinforcement learning---are fascinating entities.
They give us a chance to experiment with the processes underlying the development of human language to which we would not otherwise have access.
A perennial problem in this field, though, is that emergent languages are difficult to interpret.
The strategies emergent languages use to convey meaning do not always align with those known from human language \citep{kottur-etal-2017-natural,chaabouni2019antiefficient,kharitonov2020emergentlanguagegeneralizationacquisition}.
Yet a lack of general-purpose methods for investigating the structure of emergent communication prevents us from systematically investigating how they encode meaning.

As an essential step towards understanding emergent languages, we introduce CSAR, an algorithm for \emph{morpheme induction} on emergent language.
That is, given an input corpus of parallel data: utterances and their associated meanings, find the smallest meaningful components of utterances with their accompanying meaning.
Simply put, this task is to jointly segment utterances and align them with their meanings.
The output of this algorithm, then, is a mapping between the forms and meanings of the emergent language (example shown in \cref{fig:example}).
Furthermore, the proposed algorithm is easily applicable to almost any emergent language due to the simplicity of the input format.
In fact, the algorithm is general purpose enough to produce reasonable results in other domains, as we demonstrate with human language-based image captioning, machine translation, and word segmentation data.
Validating CSAR's performance on human language data as well as synthetic data is critical to demonstrating its effectiveness since there is no way of obtaining ground truth morphemes for emergent languages.

\begin{figure}
\centering
\begin{tabular}{ll}
  \toprule
  Form & Meaning \\
  \midrule
  3, 6 & \{not, gray\} \\
  7, 7 & \{not, blue\} \\
  32 & \{circle\} \\
  4, 5 & \{not, yellow\} \\
  6, 12, 6, 12 & \{green, or, yellow\} \\
  3, 12, 3 & \{blue, or, yellow\} \\
  \bottomrule
\end{tabular}

\caption{Example of morphemes extracted from a signalling game with pixel observations.}%
\label{fig:example}
\end{figure}

An inventory of the morphemes of an emergent language is the foundation of many further linguistic analyses.
Existing studies of compositionality \citep{korbak2020measuringnontrivialcompositionalityemergent}, word boundaries \citep{ueda2023on}, and grammar induction \citep{van-der-wal-etal-2020-grammar} could be validated and augmented with information on the morphology of emergent languages,
  and 
  new directions would also be made possible, including analyses of the morphosyntactic patterns and typological properties of emergent languages.
Ultimately, such studies form one of the pillars of emergent communication research: learning what emergent language can tell us about human language \citep{boldt2024review}.

In \cref{sec:prob} we define the task of morpheme induction and discuss related work.
\cref{sec:alg} presents the proposed algorithm, CSAR\@.
\cref{sec:val} validates CSAR's performance on procedurally generated and human language data.
\cref{sec:el} applies CSAR to emergent languages.
And we discuss the paper's findings and limitations and conclude in \cref{sec:discuss,sec:limitations,sec:conclusion}.

\paragraph{Contributions}
This work:
(1) Introduces an algorithm for inducing morphemes, applicable to a wide variety of emergent language corpora.
(2) Offers an easy-to-use Python implementation for executing the proposed algorithm on arbitrary emergent language data.
(3) Provides a first look into the morphology of emergent languages including phenomena such as polysemy and synonymy.

\section{Problem Definition}%
\label{sec:prob}
In this section we give a precise definition of the problem and the terms we will use throughout the paper.

\subsection{Task: morpheme induction}
We define the task of \emph{morpheme induction} as follows:
Given a corpus of \emph{utterances} and their corresponding \emph{complete meanings}, identify \emph{minimal, well-founded form--meaning pairs} (i.e., \emph{morphemes}) present in the corpus.
This collection of pairs is the \emph{morpheme inventory} of the corpus.

\smallskip
\begin{description}[nosep,itemindent=-1em]
  \item[form] a sequence of (form) tokens; represented as an integer sequence in emergent language.
  \item[utterance] a complete sequence of tokens produced by an agent; forms are subsequences of utterances.
  \item[meaning] a set of meaning tokens (i.e., atomic meanings grounded in the environment).
  \item[complete meaning] a meaning which represents the entire meaning of an utterance.
    \unskip\footnote{atomic meaning $\in$ meaning $\subseteq$ complete meaning}
  \item[well-founded] a form--meaning pair is well-founded when the particular form corresponds with a particular meaning.
  \item[minimal] a well-founded form--meaning pair is minimal when there is no way to decompose the pair while maintaining continuity of meanings.
\end{description}
\medskip

It is important to note that we make two assumptions about the complete meanings.
First, complete meanings are assumed to be \emph{abstracted} already, hence the reason we can represent them as a set of atomic meanings.
That is to say that the ``raw semantics'' of the utterances are already broken down into individual components of interest; this task does not entail automatically finding meaning in arbitrary data (cf.\@ clustering).
Second, since complete meanings are sets, they are not able to represent more complex phenomena that might require graph structures, for example (cf.\@ abstract meaning representations).

Additionally, we note that \emph{well-founded correspondence} is a concept subject to a variety of philosophical accounts.
Sometimes these accounts hold that the meaning is derived from either the behavior or state of mind of a language user.
Yet in this task, we only have access to a corpus, not to the language users themselves;
  thus, we employ a notion of ``well-founded correspondence'' most akin to a statistical view semantics (e.g., as in the distributional hypothesis).

\subsection{Related work}

\paragraph{Emergent language}

\citet{lipinski2024speaking} serves as the inspiration for this paper through its application of normalized pointwise mutual information to probe emergent languages for certain kinds of form--meaning relationships,
  though it stops short of providing full morpheme inventories over arbitrary data.
\citet{ueda2023on} introduces a method of form-only segment induction for emergent language based on token-level entropy patterns in utterances.

Finally, \citet{brighton} introduce methods for inducing morphemes from simulations of language evolution.
In particular, the algorithm is based on finite state transducers and the minimum description length principle.
The key difference, though, is that the FST-based method assumes a strict form--meaning correspondence that does not appear to hold in emergent languages generated by deep neural networks.
Thus, CSAR is designed to handle noise and looser form--meaning correspondence.

\paragraph{Statistical word alignment}
The task of morpheme induction resembles the task of statistical word alignment for machine translation insofar as it involves learning a mapping between two modalities.
Well-known algorithms for this task include the IBM alignment models \citep{ibm-model-1}.
While morphemes can be extracted from the alignments, the alignments themselves are not intended to represent morphemes as such.

\paragraph{Segment induction}
Segment induction is similar to morpheme induction, except that it deals only with the forms;
  these methods, then, cannot provide a mapping between form meaning because they are meaning-unaware.
Sometimes this task is called ``morpheme induction'' since the segments are supposed to correspond to morphemes, but they are not morphemes in the particular sense we use for this paper, that is: explicit form--meaning pairs.
An example of an algorithm which addresses this task is Morfessor \citep{creutz-lagus-2002-unsupervised,morfessor2} or the submissions to the SIGMORPHON 2022 Shared Task \citet{batsuren-etal-2022-sigmorphon}.
\citet{narasimhan-etal-2015-unsupervised} introduce a semi-supervised segment induction algorithm that uses semantic features to guide segmentation (viz.\@ groups of morphologically related words), although meanings are not modelled explicitly and ``word'' is not a well-defined concept for emergent languages.
The discovery of valid segments by tokenization methods based on statistics---such as BPE \citep{sennrich-etal-2016-neural,gage1994bpe} and Unigram LM \citep{kudo-2018-subword}---is largely an epiphenomenon, not a design goal.


%% file: src/algorithm.tex
\section{Algorithm}%
\label{sec:alg}

In this section we introduce the algorithm for morpheme induction: CSAR (Count, Select, Ablate, Repeat).
CSAR comprises the following steps:
\smallskip
\begin{enumerate}[nosep]
  \item Collect form and meaning candidates from the corpus.
  \item While form and meaning candidates remain.
  \begin{enumerate}[leftmargin=1.5em,nosep]
    \item Count co-occurrences of form and meaning candidates.
    \item Select form--meaning pair with the highest weight.
    \item Remove instances of the form--meaning pair from the corpus.
  \end{enumerate}
  \item Selected form--meaning pairs constitute the morpheme inventory of the corpus.
\end{enumerate}
\medskip
The code implementing CSAR as well as the experiments discussed later is available under a free and open source license at {\url{https://github.com/brendon-boldt/csar}}.

\subsection{Representation and preprocessing}%
\label{sec:alg-rep}

\paragraph{Input data}
The input data to CSAR is a parallel \emph{corpus} of utterances and their meanings.
Each record in the corpus is a tuple of form and meaning where a form is a list of (form) tokens and a meaning is a set of (meaning) tokens.

\paragraph{Candidate collection}
Given the corpus, we can identify and count the \emph{form} and \emph{meaning candidates} to produce their corresponding \emph{occurrence matrices}. 
A form candidate is any substring of form tokens under consideration for inducing morphemes.
A meaning candidate is any subset of meaning tokens under consideration for inducing morphemes.
The most straightforward approach is to simply consider every non-empty substring of forms and subset meanings, although CSAR is not constrained to this approach in theory (cf.\@ \cref{app:candidate}).

Having defined the universe of forms and meanings, we can build a binary \emph{occurrence matrix} for forms and one for meanings, where each row corresponds to a record and each entry corresponds to the presence ($1$) or absence ($0$) of a form/meaning in that record.
Thus, the form occurrence matrix has the shape
  $O_{\mathcal F}: |\mathcal R| \times |\mathcal F|$
  and the meaning matrix $O_{\mathcal M}: |\mathcal R| \times |\mathcal M|$,
  where
  $\mathcal R$ is the list of records,
  $\mathcal F$ is the set of all forms candidates,
  and $\mathcal M$ is the set of all meanings candidates.

\paragraph{Example}
If we had a simple corpus with records
  (``s'', {$\square$}),
  (``st'', {$\boxtimes$}),
  (``ct'', {$\otimes$}),
  the corresponding occurrence matrices would be:
\begin{equation}
  \mathcal O_{\mathcal F} =
    \left[
    \begin{smallmatrix}
      \cdot & \text{s} & \cdot & \cdot & \cdot \\
      \cdot & \text s & \text t & \cdot & \text{st} \\
      \text c & \cdot & \text t & \text{ct} & \cdot \\
    \end{smallmatrix}
    \right]
  \hspace{0.7em}
  \mathcal O_{\mathcal M} =
    \left[
    \begin{smallmatrix}
      \square & \cdot & \cdot & \cdot & \cdot \\
      \square & \times & \cdot & \boxtimes & \cdot \\
      \cdot & \times & \bigcirc & \cdot & \otimes \\
    \end{smallmatrix}
    \right]
    \!,
\end{equation}
where entries with value $1$'s are shown with the occurring symbols, and entries with value $0$'s with $\cdot$ for clarity.

\subsection{Main loop}

\paragraph{Weighting and co-occurrences}
Given the occurrence matrices, the next step is to compute the weights of all potential pairs.
The pair with the highest weight will be selected and added to the morpheme inventory.
The weight of a form--meaning pair is the mutual information of the binary variables representing the corresponding form and meaning.
The mutual information of a particular form--meaning pair is given by
\begin{equation}
  I(F;M) =
  \sum_{x\in F}
  \,
  \sum_{y\in M}
  p(x,y) \log_2 \frac{p(x,y)}{p(x)p(y)}
  ,
\end{equation}
where
  $F=\{f,\neg f\}$,
  $p(f)$ is the probability of $f$ appearing in a record,
  $p(\neg f)$ is the probability of $f$ not appearing,
  and the rest are defined analogously.
The key term of the mutual information expression is the joint probability between a form and a meaning, $p(f,m)$: since $f$ and $m$ are binary variables, all other joint probabilities can be computed from their joint probability and the marginal probabilities.
The joint probability can be computed by normalizing the sum of co-occurrences of given forms and meanings, namely:
\begin{equation}
  p(f,m)=
  \frac1{|\mathcal R|}
  \sum_{j=1}^{|\mathcal R|} O_{\mathcal F}[j,i_f] \wedge O_{\mathcal M}[j,i_m]
\end{equation}
where $i_f$ and $i_m$ are the indices of $f$ and $m$ in their respective matrices.
More succinctly, co-occurrences can be computed with matrix multiplications, yielding
\begin{equation}
  p(f,m) =
  \frac1{|\mathcal R|}\cdot\left(O_{\mathcal F}^\top O_{\mathcal M}\right)[i_f, i_m]
\end{equation}
Other weighting methods were explored including joint probabilities, pointwise mutual information, and normalized pointwise mutual information, though mutual information was found to perform best empirically.

The above weighting function results in ties which we break with the following heuristics:
  (1) higher initial weight,
  (2) fewer selected pairs with this form,
  (3) larger form size,
  and (4) smaller meaning size.

\paragraph{Remove pair from corpus}
The final step of the algorithm's main loop is ablating the pair from the corpus.
That is, once we select a form--meaning pair, we want to remove all co-occurrences of the form and meaning in order to determine what form--meaning correspondences remain to be explained.
For example, after ablating the pair (``t'', {$\times$}), the corpus from above would comprise
  (``s'', {$\square$}),
  (``s'', {$\square$}),
  and (``c'', {$\Circle$});
  the occurrence matrices would then be updated to reflect this.
In cases where ablating a pair is ambiguous, we apply a heuristic (see \Cref{app:ambig-app}).

\paragraph{Repeating and stopping}
After ablating the selected form--meaning pair, the algorithm repeats the main loop, beginning again at the weight-computation step (with the updated occurrence matrices).
The one difference is that---in subsequent weight computations---the weight of a pair cannot go up, preventing spurious correlations from arising in later steps.

This loop continues until form or meaning occurrences are exhausted or some other criterion is met (e.g., time limit, inventory size limit).
In this way, CSAR is an ``anytime'' algorithm since it can be stopped after an arbitrary number of iterations and still produce a sensible result.
This is because the most heavily weighted morphemes can be considered the highest \emph{confidence} morphemes, meaning that stopping the algorithm before completion will only leave out the lowest confidence morphemes.

\subsection{Implementation}
The implementation of CSAR introduced in this paper is written in Python making use of sparse matrices from \texttt{scipy} \citep[BSD 3-Clause license]{scipy} and JIT compilation with \texttt{numba} \citep[BSD 2-Clause license]{numba} to speed up execution.
CSAR is conceptually simple. Most of the implementation complexity lies in efficiently handling the occurrence matrices, especially when removing a form--meaning pair from the corpus.
For example, the co-occurrence matrix has the shape $|\mathcal F| \times |\mathcal M|$ which is massive considering that $\mathcal F$ and $\mathcal M$ are already accounting for the universes of all possible forms and meanings in the corpus.
Nevertheless, there are a wide range of heuristics that can be applied to greatly speed up execution while maintaining performance (see \cref{app:opt}).


%% file: src/validation.tex
\section{Empirical Validation}%
\label{sec:val}
To validate the ability of CSAR to find well-founded morpheme inventories, we test it against procedurally generated datasets as well as human languages.
Since we do not have access to ground truth morphemes for emergent languages, we gauge the effectiveness of CSAR's morpheme induction in the next best way: by testing its performance in these adjacent domains.
Procedurally generated datasets (described in \cref{sec:data-gen}) both give us access to the ``ground truth'' morphemes and allow us to vary particular facets of the languages.
Having ground truth morphemes allows us to quantitatively evaluate CSAR against baseline methods (\cref{sec:baselines}).
Fine-grained control over the facets of the languages permits us to identify particular phenomena that are challenging for CSAR to induce correctly (\cref{sec:error-ana}).
We also test CSAR against human language data (\cref{sec:human-lang}) in order to give a qualitative sense of the effectiveness of the algorithm.

\subsection{Procedural datasets}
\unskip\label{sec:data-gen}

The dataset-generating procedure has the following basic structure:
  (1) Meanings are sampled according to some structure (viz.\@ a fixed attribute--value vector).
  (2) An utterance is produced from this meaning according to a mapping of meaning components to form components.
  (3) The form--meaning pairs that were used to generate the utterance are added to the set of ground truth morphemes.
In the basic case, for example, each particular attribute and value is associated with a unique sequence of tokens which are concatenated to form an utterance, creating a one-to-one mapping from meanings to forms.

\paragraph{Variations}
Such languages are trivial to induce morphemes from, so we introduce the following variations to produce more complex datasets:
\smallskip
\begin{description}[nosep,itemindent=-1em]
  \item[Synonymy] Multiple forms may correspond to the same meaning.
  \item[Polysemy] Multiple meanings may correspond to the same form.
  \item[Multi-token forms] A form may comprise more than one token, possibly overlapping with other forms.
  \item[Vocab size] Number of unique tokens.
  \item[Sparse meanings] Meanings occur independently of each other with no additional structure (i.e., not structured as attribute--value pairs).
  \item[Distribution imbalance] Meanings are sampled from non-uniform distributions.
  \item[Dataset size] Number of records in the dataset.
  \item[Number of meanings] Total number of meanings (e.g., varying number of attributes and values).
  \item[Noise forms] Form tokens not corresponding to any meanings are added.
  \item[Shuffle form] Inter-form order is varied randomly (while maintaining intra-form order).
  \item[Non-compositionality] A given form may correspond to multiple meanings simultaneously.
\end{description}
\medskip
For the following analyses, we report values for a collection of procedural datasets built from the Cartesian product of two values for each of the above variations (one where the variation is inactive and one where it is).
See \cref{app:synth-hparams} for details.

\paragraph{Evaluation metric}
We use $F_1$ score (harmonic mean of precision and recall) to assess the quality of an induced morpheme inventory given the ground truth inventory.
We define precision as
\begin{equation}
  \frac1{|\mathcal I|} \sum_{i \in \mathcal I} \max_{g\in\mathcal G} s(i, g) ,
  \label{eq:precision}
\end{equation}
where
  $\mathcal I$ is the set of induced morphemes,
  $\mathcal G$ is the set of ground truth morphemes,
  and
  $s$ is the similarity function
  \unskip.
For exact $F_1$, the similarity function is $1$ if the morphemes are identical and $0$ otherwise.
In fuzzy $F_1$, the similarity function is the minimum of form similarity (normalized insertion--deletion ratio\footnote{$1 - (\text{insertions} + \text{deletions}) / (|s_1|+|s_2|)$}) and meaning similarity (Jaccard index).
Recall is defined similarly to precision except that the roles of $\mathcal{I}$ and $\mathcal{G}$ from \cref{eq:precision} are reversed.


\subsection{Comparison with baselines}
\unskip\label{sec:baselines}

\begin{figure}
  \centering
  \input{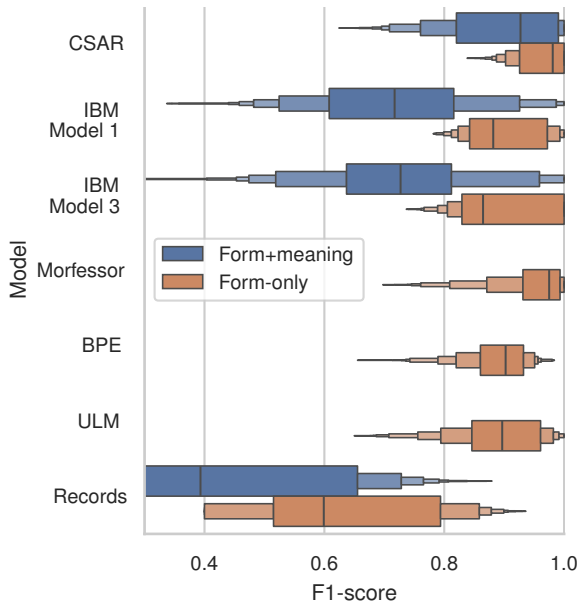}
  \caption{Fuzzy $F_1$ scores for CSAR and baseline methods across procedural datasets.
    Results reported for form--meaning inventories and form-only inventories.}
  \unskip\label{fig:baseline}
\end{figure}

Below we describe the baseline methods we use for comparison.
\smallskip
\begin{description}[nosep,itemindent=-1em]
  \item[IBM Model 1]
    Simple expectation-maximization approach to machine translation primarily through aligning words in a sentence-parallel corpus. \citep{ibm-model-1}
  \item[IBM Model 3]
    Built on top of the IBM Model 1 to handle phenomena such as allowing a form to align to no meaning.
  \item[Morfessor]
    A form-only segmentation algorithm built to handle human language; also uses an EM algorithm.
  \item[Byte pair encoding]
    A greedy form-only tokenization method which recursively merges frequently occurring pairs of tokens.
    Vocabulary size is selected according to a simple heuristic (see \Cref{app:vocab-size}).
    \citep{gage1994bpe,sennrich-etal-2016-neural}
  \item[Unigram LM]
    An EM-based form-only tokenization method which starts with a large vocabulary and iteratively removes tokens contributing least to the likelihood of the data.
    Vocabulary size is selected according to a simple heuristic (see \Cref{app:vocab-size}).
    \citep{kudo-2018-subword}
  \item[Record]
    A trivial baseline where the inventory is just the set of all records.
\end{description}
\medskip
For the baseline methods which do not handle meanings and only produce forms, we report the form-only $F_1$ score (i.e., $s(i,g)$ only takes the form into account), though CSAR and IBM models still have access to meanings.
For form-only metrics, we exclude datasets which include noise forms as form-only methods cannot identify which forms are noise.

\paragraph{Results}
The results of CSAR and the baselines on the procedural datasets are presented in \cref{fig:baseline},
  which shows the distributions of mean scores for each hyperparameter setting for the procedural datasets. Each setting was repeated over $3$ random seeds. Additional results are given in \cref{app:proc-table}.
For inducing full morphemes (form and meaning), CSAR performs the best by a large margin over the baselines (and even greater margin when considering exact $F_1$).
The IBM alignment models perform better than the trivial record-based baseline but still perform noticeably worse than CSAR\@.
While CSAR yields roughly equal precision and recall, the IBM models' precision is lower than their recall suggesting that they are more prone to inducing spurious morphemes than CSAR\@.

When evaluating the forms only, we find that CSAR is the best method with Morfessor exhibiting comparable performance.
The IBM alignment models exhibit roughly the same performance as the tokenization methods (BPE and Unigram LM).
As with the full morpheme results, CSAR is the only method to achieve comparable precision and recall with all other baselines having precisions lower than their recalls.

\subsection{Error Analysis}
\unskip\label{sec:error-ana}
For the most part, the errors CSAR makes are ``edit errors'': identifying a correct morpheme but adding or removing a form or meaning token.
This is reflected in the near-parity between precision and recall.
This is in contrast to the baseline methods which are more prone to inducing too many morphemes, leading to lower precision.

Generally speaking, as more variations are added to a dataset, the performance degrades further.
In particular, CSAR's performance decreases the most with small corpus sizes, overlapping multi-token forms, and non-compositional mappings.
On the other hand, using sparse meanings, shuffling the forms, and using a non-uniform meaning distribution have relatively little effect.

\begin{figure}
\centering
\begin{tabular}{lp{10em}}
  \toprule
  Dataset & Induced Morpheme \\
  \midrule
  \multirow2*{Morphology}     & (``ed\$'', \{PAST\}) \\
                              & (``\,'\,'', \{POSSESSIVE\}) \\
  \midrule
  \multirow3*{Image captions} & (``stop sign'', \{stop sign\}) \\
                              & (``woman'', \{person\}) \\
                              & (``skier'', \{person, skis\}) \\
  \midrule
  \multirow2*{Translation}    & (``Member States'', \{Mitgliedstaaten\}) \\
  \bottomrule
\end{tabular}
\caption{Examples of morphemes induced from various human language datasets and tasks.}%
\label{fig:human}
\end{figure}

\subsection{Human language data}
\unskip\label{sec:human-lang}

In this section we discuss the results of running CSAR on three different human language datasets.
While these datasets are not the intended domain of CSAR---and CSAR is certainly not the best algorithm for the tasks---the point of these experiments is to demonstrate the general effectiveness of the algorithm qualitatively (examples shown in \cref{fig:human}).
Since these datasets are larger, we employed heuristic optimizations to CSAR to reduce their runtime (described in \cref{app:opt}).
The top $100$ induced morphemes for each human language dataset are given in \cref{app:human-language}.

\paragraph{Morpho Challenge}
The first human language dataset we use is from the Morpho Challenge \citep{kurimo-etal-2010-morpho}.
This dataset is a human language approximation of the task of morpheme induction for emergent language.
Concretely, the utterances are single English words, divided up at the character level, while the meanings are the constituent morphemes.

CSAR is able to recover a wide variety of morphemes including:
  roots like (``\^{}fire'', \{fire\}),
  prefixes like (``\^{}re'', \{re-\}),
  suffixes like (``ed\$'', \{PAST\}),
  and other affixes like (``\,'\,'', \{POSSESSIVE\}).
While the vast majority of morphemes CSAR induces are accurate, a handful of the lowest-weighted morphemes are spurious (e.g., (``s\$'', \{boy\})) likely due to inaccurate decoding earlier in the process (i.e., part of the true form for a given meaning was included in a prior meaning).

\paragraph{Image captions}
The next dataset we employ is the MS COCO dataset \citep[CC BY 4.0]{lin2015microsoftcococommonobjects}.
In particular, we take the image captions to be the utterances, treating words as atomic units, and the meaning to be the labeled objects in the image (e.g., person, cat).

The bulk of highest weighted induced morphemes are direct equivalents of the objects they describe (e.g., (``cat'', \{cat\})).
We find instances of synonymy (e.g., (``bicycle'', \{bicycle\}) and (``bike'', \{bicycle\})) as well as polysemy (e.g., (``animals'', \{cow\}) and (``animals'', \{sheep\})).
Finally, we also observe compound forms like (``stop sign'', \{stop sign\}) as well as compound meanings such as (``skier'', \{person, skis\}).
As we go beyond the top $100$ or so, the associations between forms and meanings remain reasonable but become looser such as (``bride'', \{dining table, tie\}) or (``sink'', \{toothbrush\}).

\paragraph{Machine translation}
For machine translation, we use the WMT16 dataset and the English--German split, in particular \citep{bojar-EtAl:2016:WMT1}.
In this case, the English text is considered to be the utterance and the German text to be the meaning, with words being the atomic units on both sides.

As with the image caption results, the bulk of induced morphemes are direct equivalents (e.g., (``and'', \{und\})).
Beyond these simple one-to-one mappings, CSAR induces the polysemic relationship (``the'', \{der\}) and (``the'', \{die\}).
Finally, CSAR also picks up on multi-token forms like (``Member States'', \{Mitgliedstaaten\}).

%% file: src/ec.tex
\begin{table*}
\centering
\input{assets/ec-table}
\caption{Morpheme inventory metrics (described in \cref{sec:ec-quant}) across various emergent languages. (AV: attribute--value, SW: ShapeWorld, Inv.: Inventory)}%
\label{tab:ec-quant}
\end{table*}

\section{Analysis of Emergent Languages}%
\label{sec:el}
\subsection{Datasets}
We apply CSAR to two different signalling game environments: one with vector-based observations and one with image-based observations.

\paragraph{Vector observations}
In the vector observation signalling game the agents directly observe one-hot vectors which directly correspond to the information to be communicated \citep[MIT license]{egg}.
Specifically, we use two variants:
  (1) the standard attribute--value setting where each of $4$ attributes can take on $4$ distinct values
  and (2) the ``sparse'' setting where there are $8$ binary attributes and only attributes which are ``true'' are included in the meanings given to CSAR\@.
Hyperparameters for both environments are given in \cref{app:el-hparams}.

\paragraph{ShapeWorld observations}
The second environment is introduced by \citet[MIT license]{mu2021general} with the following differences:
  (1) observations are images,
  and (2) employs variations which increase the level of abstraction of the game to encourage generalization.
First, this environment uses the ShapeWorld tool for generating observations \citep{kuhnle2017shapeworldnewtest};
  namely, underlying concepts are particular shapes (e.g., red square) while the observations passed to the agents in the signalling game are pixel-based images.
Second, \citet{mu2021general} provide three variants with increasing levels of abstraction:
  (1) \emph{reference}: the sender indicates a single image,
  (2) \emph{set reference}: the sender indicates a set of images with a common attribute,
  and (3) \emph{concept}: as in \emph{set reference} but the receiver's observations are different instances sharing the common attribute (referenced in \cref{fig:example}).

\subsection{Metrics}%
\label{sec:ec-quant}
We present the following metrics to give analyze the morpheme inventories induced from the emergent language data:
\smallskip
\begin{description}[nosep,itemindent=-1em]
  \item[Inventory size] Number of morphemes in the inventory.
  \item[Inventory entropy] Entropy (in bits) of the morphemes according to their prevalence.
  \item[Synonymy] Entropy across forms for a given meaning.
  \item[Polysemy] Entropy across meanings for a given form.
  \item[Form size] Mean number of tokens in a form.
  \item[Meaning size] Mean number of tokens in a meaning.
  \item[Topographic similarity] Correlation (Spearman's $\rho$) between distances in the utterance space and complete meaning space \citep{brighton2006toposim,lazaridou2018EmergenceOL}.
\end{description}
\smallskip
With the exception of inventory size and toposim, the above metrics are weighted by \emph{prevalence} which is the proportion of records from which the morpheme was ablated.

\subsection{Results}
\Cref{tab:ec-quant} shows the results (induced morphemes from each emergent language are given in \cref{app:ec-inv}).
Looking at form size, while the forms of morphemes do tend towards smaller values, many comprise more than one token, suggesting that assuming that each token can be analyzed as a word or independent unit of meaning is not a safe assumption.
Addressing the mapping between forms and meanings, we see that synonymy (forms per meaning) is higher than polysemy (meanings per form).
The fact that there is a higher degree of synonymy than polysemy makes sense insofar as the optimization penalizes ambiguity (polysemy) while it does not penalize merely inefficient encoding (synonymy).
This is concordant with finding such as \citet{chaabouni2019antiefficient} which finds that emergent languages, in the absence of addition pressures, do not develop efficient encoding schemes.

\paragraph{Compositionality}
The meaning size metric, in particular, is interesting insofar as it relates to compositionality.
In the simplest case of compositionality, morphemes comprise singleton meanings which can be combined to form compound meanings.
More holistic languages, on the other hand, assign multiple atomic meanings per morpheme resulting in in larger meaning sizes.
The fact that the emergent languages tend towards a meaning size of $1$ suggests a non-trivial degree of compositionality under this interpretation.
Yet when we compare meaning sizes values to topographic similarity values computed across records (i.e., not involving CSAR), we find that there is no obvious correlation between toposim values and meaning sizes.
This could be due to the fact that individual form tokens could have ``partial meanings'' and need to be combined to comprise an atomic meaning.
Although our sample size is too small to make any definitive claims.

%% file: assets/ec-table.tex
\begin{tabular}{lrrrrrrr}
\toprule
 & $|\text{Inv.}|$ & Inv.\@ $H$ & $|\text{Form}|$ & $|\text{Meaning}|$ & Synonymy & Polysemy & Toposim \\
\midrule
Vector, AV & 223 & 6.81 & 3.07 & 1.37 & 1.52 & 0.58 & 0.35 \\
Vector, sparse & 156 & 6.09 & 2.08 & 1.55 & 1.91 & 0.62 & 0.39 \\
SW, ref & 1124 & 6.52 & 1.76 & 1.01 & 2.99 & 1.64 & 0.04 \\
SW, setref & 396 & 6.14 & 1.54 & 1.38 & 1.43 & 0.74 & 0.15 \\
SW, concept & 351 & 5.86 & 1.89 & 1.43 & 1.04 & 0.95 & 0.17 \\
\bottomrule
\end{tabular}

%% file: src/outro.tex
\section{Discussion}%
\label{sec:discuss}

Due to CSAR's strong performance and easy application to a wide variety of emergent language environments, it would be a valuable addition to the standard toolkit of emergent language analyses.
In particular, it helps fill a gap of environment-agnostic methods for interpreting the ways that emergent languages convey meaning---a perennial question in the field. 
Down the road, this opens up research questions concerning the evolution of meaning in emergent language, such as those discussed in \citet{brighton}, but with the ability to deal with the larger scale and particular difficulties of \emph{deep learning-based} emergent communication.

Furthermore, morpheme inventories are a foundation for higher-level linguistic analyses of emergent language like inducing their syntactic structure.
To skip the morpheme induction step would be comparable to attempting to understand the grammatical role of the letter \emph{C} in English.
Such analyses of the syntax of emergent language and beyond are critical to understanding how emergent and human language are similar and how they are different.


\section{Conclusion}%
\label{sec:conclusion}

CSAR presents a strong platform for investigating the morphology of emergent language, demonstrating the ability to find minimal form--meaning pairs in both procedural and human language data.
Given the morpheme inventory of an emergent languages we can not only analyze phenomena like synonymy and polysemy but also the typological features of emergent languages, determining which human languages they most closely resemble, if they resemble any.
Such a study of morphology forms the foundation for the more general study of the linguistic features of emergent language and unlocks the door to the insights they can provide us about human language.

\section{Limitations}%
\label{sec:limitations}

\paragraph{Greed is not always good}
While the greediness of CSAR does simplify induction (conceptually and implementation-wise), improve runtime, and provide good partial inventories,
  it suffers from the same limitation inherent to greedy algorithms: it can get trapped in local optima.
For example, it is possible to construct corpora for which a greedy approach is ``misled'' since certain heuristics require revision based on information encountered later in the induction process (e.g., preferring smaller versus larger forms).

We did consider non-greedy approaches to morpheme induction but ultimately decided not to pursue them in this work because (1) the greedy approach itself demonstrated strong performance and (2) initial attempts at non-greedy approaches (e.g., tree search) yielded intractable runtimes.
For example, an error due to greediness might select morpheme $B$ before morpheme $A$ because $B$ had a higher weight while $A$ was ultimately correct.
To select $A$ instead of $B$, the morpheme candidates would have to be reordered which, without an efficient way to propose these order, worsens the time complexity from $O(n^c)$ to $O(n!)$.
Related algorithms use iterative approaches (IBM models and Morfessor) or search \citep{brighton} to avoid the local minima that trap greedy approaches.
Future work could incorporate such methods to improve upon the performance of CSAR for morpheme induction.

\paragraph{Limited emergent language data}
The other limitation of this paper relates to the type and breadth of emergent language data.
In terms of type, since we do not have ground truth morpheme inventories for emergent language, we cannot directly evaluate CSAR's performance on the target domain (hence the validation with procedurally generated and human languages).
In terms of breadth, without a larger and more representative sample of more systematically generated data we are unable to make definitive claims about the patterns and trends of morpheme inventories in emergent languages.

%% file: src/appendix.tex
\appendix

\section{Algorithm}%
\label{app:alg}

\subsection{Candidate generation}%
\label{app:candidate}

For simplicity's sake (and inductive bias), we limit the candidate generation functions to all non-empty substrings for forms and all non-empty subsets for meanings.
Nevertheless, we could extend form candidate generation to non-contiguous forms to detect non-concatenative morphology (e.g., the form ``\mbox{x.z}'' matching ``\mbox{xyz}'' and ``\mbox{xwz}'').
In fact, we could could use arbitrary regular expressions to represent forms (or meanings) such as ``\^{}..x'' or ``x+'' to represent absolute position and optional repetitions, respectively.
We could consider empty forms and empty meanings to explicitly identify forms and meanings which do not have mappings (as opposed to implicitly not including them in the morphology).

Of course, part of the difficulty of extending the complexity of the candidate generation is that it expands the already (sometimes intractably) large search space.
One method of making this tractable, though, is adding heuristics that determine which form candidates should be considered rather than considering every possible candidate.

\subsection{Ambiguous pair application}%
\label{app:ambig-app}

In some cases of applying a morpheme to record in the dataset, there are multiple applications possible.
Say we have the utterance ``x y z x y'' meaning $\{A,B\}$ and we want to apply the morpheme (``x y'', $\{A\}$).
The form matches two substrings in the utterance, so there are two possible ways to apply the morpheme.
As a heuristic for selecting the best application, CSAR break ties by selecting the substring least likely to be a morpheme (as determined by the morpheme weights).
Going back to the above example, if it is the case the morpheme (``z x y'', $\{B\}$) has a higher weight than (``x y z'', $\{B\}$), then CSAR will apply (``x y'', $\{A\}$) to the first instance of ``x y'' instead of the second.

This search can be very computationally expensive since it can entail going through a large number of morpheme candidates.
Thus for the experiments with human language data, we do not perform this search and select the best form quasirandomly.

\subsection{Heuristic optimizations}%
\label{app:opt}

Below we include a summary of heuristic optimizations available in CSAR\@:
\smallskip
\begin{description}[nosep,itemindent=-1em]
  \item[max input records] Only consider a certain number of records from the input data; $20\,000$ for machine translation, image captions, and ShapeWorld.
  \item[max inventory size] Stop after inducing a certain number of morphemes; $300$ for image captions and machine translation settings.
  \item[\textit{n}-gram semantics] Treat complete meanings as ordered and generate meaning candidates identically to forms (i.e., as $n$-grams); used for machine translation data where the ``meanings'' are sentences.
  \item[max form/meaning size] Only consider form/meaning candidates up to a certain size; $3$ for machine translation (form and meaning) and image captions (form only), $2$ for image captions meaning.
  \item[no search best form] When ablating a form with multiple matches in an utterance, do not search for best form, simply choose it randomly; no search for image captions and machine translation.
  \item[form/meaning vocabulary size] Only consider the most common form/meaning candidates; $100\,000$ for image captions and machine translation.
  \item[token vocabulary size] Only consider the most common form/meaning tokens and ignore an form meaning candidates which contain an unknown token; $1000$ for image captions and $500$ for machine translation.
  \item[co-occurrence threshold] Zero out any co-occurrences which fall below a certain threshold (e.g., if a form and meaning candidate only occur once, treat it as never co-occurring); $1$ for ShapeWorld, $10$ for image captions, and $100$ for machine translation.
\end{description}

\section{Empirical Validation}%
\label{app:emp-val}

\subsection{Procedural dataset hyperparameters}%
\label{app:synth-hparams}

The following hyperparameters were used for generating the procedural datasets.
Each dataset uses $4$ attributes and $4$ values except for the sparse setting which uses $8$ independent values.
\begin{description}[nosep,itemindent=-1em]
  \item[Synonymy] $\{1,3\}$; forms per meaning
  \item[Polysemy] $\{0, 0.15\}$; proportion of meanings mapped to an already-used form
  \item[Multi-token forms] $\{\{1\}, \{1,2,3,4\}\}$; possible tokens per form
  \item[Vocab size] $\{10, 50\}$; only applies to non-unity multi-token forms
  \item[Sparse meanings] $\{\text{true}, \text{false}\}$
  \item[Distribution imbalance]  $\{\text{true}, \text{false}\}$; non-uniform distribution is based on the ramp function, i.e., probability of given value for an attribute is proportional to its $\text{index} +1$.
  \item[Dataset size] $\{50, 500\}$
  \item[Noise forms] $\{0, 0.5\}$; $1-p$ of parameter of geometric distribution
  \item[Shuffle form] $\{\text{true}, \text{false}\}$
  \item[Non-compositionality] $\{\text{true}, \text{false}\}$
  \item[Random seeds] $3$ per hyperparameter setting
\end{description}
Non-unity polysemy and synonymy rates for the non-compositional dataset implementation were not implemented and are excluded from the above grid.

\subsection{Tokenizer vocabulary size}%
\label{app:vocab-size}

The heuristic for the tokenizer vocabulary size is as follows:
\begin{align}
  |V| &= \left\lfloor
    \frac{|\mathcal T_\text{meaning}|}{|\mathcal R|} \sum_{r\in\mathcal R} \frac{|r_\text{form}|}{|r_\text{meaning}|}
  \right\rfloor
    + |\mathcal T_\text{form}|
  ,
\end{align}
where
  $\mathcal T_\text{meaning}$ is the set of all meaning tokens in the dataset (likewise for $\mathcal T_\text{form}$),
  $\mathcal R$ is the multiset of records in dataset,
  $r_\text{form}$ is the particular form (utterance) for an individual record (likewise for $r_\text{meaning}$.
This heuristic can be interpreted as the mean form tokens per meaning tokens times the number unique meaning tokens added to the number of unique form tokens (since each of them will automatically be included in the vocabulary).

\subsection{Additional procedural dataset results}%
\label{app:proc-table}

\begin{figure}
\centering
\input{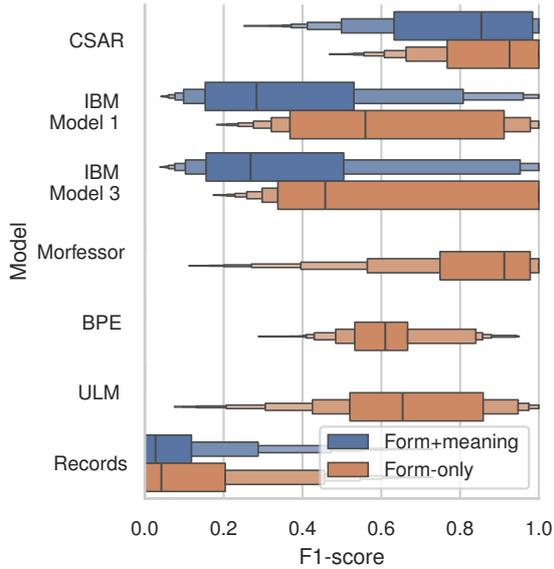}
\caption{Exact $F_1$ scores of baseline methods on the procedural datasets}%
\label{fig:baseline-exact}
\end{figure}

\Cref{tab:proc-all} shows all results of baseline methods on the procedural datasets.
\Cref{fig:baseline-exact} visualizes the results of the baseline methods with exact $F_1$ score.

\begin{table*}
\centering
\input{assets/proc-table}
\caption{Results of baseline methods on the procedural datasets.}%
\label{tab:proc-all}
\end{table*}

\section{Analysis of Emergent Languages}

\subsection{Emergent language hyperparameters}%
\label{app:el-hparams}

The following hyperparameters were used for the vector observation environment:
\smallskip
\begin{description}[nosep,itemindent=-1em]
  \item[\textit{\textbf{n}} values] $4$, $2$ (sparse)
  \item[\textit{\textbf{n}} attributes] $4$, $8$ (sparse)
  \item[\textit{\textbf{n}} distractors] $3$
  \item[vocab size] $32$
  \item[max sequence length] $10$
  \item[dataset size (CSAR input)] $10\,000$ records
\end{description}
\medskip

The ShapeWorld observation environment uses the following hyperparameters
\smallskip
\begin{description}[nosep,itemindent=-1em]
  \item[observations] $5$ shapes, $6$ colors, $3$ operators (and, or, not); \emph{and} or \emph{or} may only be used once
  \item[\textit{n} examples] $20$ total; $10$ correct targets, $10$ distractors
  \item[vocab size] $32$
  \item[max sequence length] $8$
  \item[dataset size (CSAR input)] $20\,000$ records
\end{description}
\medskip

Both environments had any beginning-of-sentence and end-of-sentence tokens removed before being fed into CSAR\@.
Running the above experiments requires about $25$ GPU-hours on NVIDIA GeForce RTX 2080Ti.

\section{Morfessor Results on Emergent Language}
\unskip\label{sec:morf-ec}
\begin{table}
  \centering
  \begin{tabular}{lrr}
  \toprule
                 & $|\text{Inv.}|$ & $|\text{Form}|$ \\
  \midrule
  Vector, AV     &            $94$ & $3.59$ \\
  Vector, sparse &           $126$ & $3.51$ \\
  SW, ref        &          $2898$ & $6.93$ \\
  SW, setref     &          $2920$ & $8.13$ \\
  SW, concept    &          $1565$ & $7.77$ \\
  \bottomrule
  \end{tabular}
  \caption{Metrics for form-only morpheme inventories generated by Morfessor across various emergent languages.}
  \unskip\label{tab:morf-ec}
\end{table}

In \cref{tab:morf-ec} we show the results of running Morfessor on various emergent language corpora.
Compared to the metrics for CSAR's output on the same corpora (\cref{tab:ec-quant}), Morfessor's results do not match or even differ consistently (although Morfessor's forms do not have prevalence weighting like CSAR's).
For the vector environments, Morfessor yields smaller inventories than CSAR yet larger inventories for ShapeWorld.
Form lengths are similar for the vector environment, but for ShapeWorld, CSAR yields shorter forms than the vector environment while Morfessor yields much longer forms.
Since we do not have ground truth morphemes for these emergent language corpora, we cannot definitively say one algorithm has performed better than the other.
Yet Morfessor here is at a disadvantage here as it is not able to use the meanings of the utterances to guide its induction.

\section{Morpheme Inventories}
Top $100$ morphemes induced by CSAR from human and emergent language datasets.

\subsection{Human languages}%
\label{app:human-language}

\paragraph{Morpho Challenge}
\input{assets/morpho-challenge}
\paragraph{Image captions}
\input{assets/coco}
\paragraph{Machine translation}
\input{assets/mt}

\subsection{Emergent languages}%
\label{app:ec-inv}

\paragraph{Vector, attribute--value}
Note that meanings are in the format \emph{\mbox{attribute\_value}} meaning that 1\_2 means the $1$\textsuperscript{st} attribute has value $2$.

\medskip
\noindent
{\input{assets/ec-vector-av}}

\paragraph{Vector, bag of meanings}
{\input{assets/ec-vector-sparse}}
\paragraph{Shapeworld, reference}
{\input{assets/ec-shapeworld-ref}}
\paragraph{Shapeworld, set reference}
{\input{assets/ec-shapeworld-setref}}
\paragraph{Shapeworld, concept}
{\input{assets/ec-shapeworld-concept}}

%% file: assets/proc-table.tex
\begin{tabular}{lrrrrrrr}
\toprule
 & CSAR & IBM Model 1 & IBM Model 3 & Morfessor & BPE & ULM & Records \\
\midrule
Exact $F_1$, form & 0.868 & 0.616 & 0.595 & 0.827 & 0.624 & 0.670 & 0.133 \\
Fuzzy $F_1$, form & 0.960 & 0.899 & 0.893 & 0.949 & 0.890 & 0.891 & 0.637 \\
Fuzzy prec., form & 0.954 & 0.855 & 0.850 & 0.933 & 0.852 & 0.853 & 0.597 \\
Fuzzy recall, form & 0.967 & 0.952 & 0.946 & 0.967 & 0.934 & 0.938 & 0.701 \\
Exact $F_1$ & 0.788 & 0.375 & 0.379 & 0.000 & 0.000 & 0.000 & 0.101 \\
Fuzzy $F_1$ & 0.899 & 0.721 & 0.726 & 0.000 & 0.000 & 0.000 & 0.441 \\
Fuzzy prec. & 0.881 & 0.641 & 0.640 & 0.000 & 0.000 & 0.000 & 0.390 \\
Fuzzy recall & 0.921 & 0.855 & 0.866 & 0.000 & 0.000 & 0.000 & 0.543 \\
\bottomrule
\end{tabular}

%% file: assets/morpho-challenge.tex
(``''', \{+GEN\})
(``ing\$'', \{+PCP1\})
(``ed\$'', \{+PAST\})
(``s'', \{+PL\})
(``er'', \{er\_s\})
(``ly\$'', \{ly\_s\})
(``s\$'', \{+3SG\})
(``ist'', \{ist\_s\})
(``iz'', \{ize\_s\})
(``ness'', \{ness\_s\})
(``ion'', \{ion\_s\})
(``\^{}re'', \{re\_p\})
(``\^{}de'', \{de\_p\})
(``ation'', \{ation\_s\})
(``est\$'', \{+SUP\})
(``\^{}un'', \{un\_p\})
(``less'', \{less\_s\})
(``ful'', \{ful\_s\})
(``\^{}mis'', \{mis\_p\})
(``head'', \{head\_N\})
(``way'', \{way\_N\})
(``ment'', \{ment\_s\})
(``al'', \{al\_s\})
(``it'', \{ity\_s\})
(``\^{}fire'', \{fire\_N\})
(``ency\$'', \{ency\_s\})
(``hook'', \{hook\_N\})
(``ish\$'', \{ish\_s\})
(``mind'', \{mind\_N\})
(``\^{}in'', \{in\_p\})
(``at'', \{ate\_s\})
(``if'', \{ify\_s\})
(``able\$'', \{able\_s\})
(``ically\$'', \{ally\_s\})
(``\^{}inter'', \{inter\_p\})
(``\^{}photo'', \{photo\_p\})
(``\^{}hand'', \{hand\_N\})
(``\^{}scho'', \{school\_N\})
(``house'', \{house\_N\})
(``ical\$'', \{ical\_s\})
(``hold'', \{hold\_V\})
(``long'', \{long\_A\})
(``work'', \{work\_V\})
(``up'', \{up\_B\})
(``ag'', \{age\_s\})
(``ant'', \{ant\_s\})
(``ib'', \{ible\_s\})
(``line'', \{line\_N\})
(``ed\$'', \{ed\_s\})
(``er\$'', \{+CMP\})
(``\^{}over'', \{over\_p\})
(``\^{}dis'', \{dis\_p\})
(``\^{}sea'', \{sea\_N\})
(``\^{}im'', \{im\_p\})
(``or'', \{or\_s\})
(``pos'', \{pose\_V\})
(``ence'', \{ence\_s\})
(``\^{}cardinal'', \{cardinal\_A\})
(``\^{}rational'', \{rational\_A\})
(``\^{}shoplift'', \{shop\_N\})
(``conciliat'', \{conciliate\_V\})
(``\^{}manicur'', \{manicure\_N\})
(``\^{}predict'', \{predict\_V\})
(``dressing'', \{dressing\_V\})
(``\^{}buffet'', \{buffet\_V\})
(``\^{}crimin'', \{crime\_N\})
(``\^{}entitl'', \{entitle\_V\})
(``\^{}frivol'', \{frivolous\_A\})
(``\^{}heartb'', \{heart\_N\})
(``\^{}maroon'', \{maroon\_A\})
(``\^{}ribald'', \{ribald\_A\})
(``\^{}spread'', \{spread\_V\})
(``\^{}squeak'', \{squeak\_V\})
(``\^{}squint'', \{squint\_V\})
(``\^{}statue'', \{statue\_N\})
(``\^{}summar'', \{summary\_A\})
(``whisper'', \{whisper\_V\})
(``\^{}blink'', \{blink\_V\})
(``\^{}carri'', \{carry\_V\})
(``\^{}cheer'', \{cheer\_V\})
(``\^{}four-'', \{four\_Q\})
(``\^{}hitch'', \{hitch\_V\})
(``\^{}louvr'', \{louvre\_N\})
(``\^{}muzzl'', \{muzzle\_N\})
(``\^{}nihil'', \{nihilism\_N\})
(``\^{}tooth'', \{tooth\_N\})
(``\^{}waist'', \{waist\_N\})
(``guard\$'', \{guard\_N\})
(``\^{}bull'', \{bull\_N\})
(``\^{}rail'', \{rail\_V\})
(``\^{}seri'', \{series\_N\})
(``\^{}test'', \{test\_N\})
(``\^{}two-'', \{two\_Q\})
(``ance\$'', \{ance\_s\})
(``board'', \{board\_N\})
(``chain'', \{chain\_N\})
(``eroom'', \{room\_N\})
(``grand'', \{grand\_A\})
(``order'', \{order\_V\})
(``power'', \{power\_N\})

%% file: assets/coco.tex
(``tennis'', \{tennis racket\})
(``cat'', \{cat\})
(``train'', \{train\})
(``dog'', \{dog\})
(``pizza'', \{pizza\})
(``toilet'', \{toilet\})
(``man'', \{person\})
(``bus'', \{bus\})
(``clock'', \{clock\})
(``baseball'', \{baseball glove\})
(``frisbee'', \{frisbee\})
(``bed'', \{bed\})
(``horse'', \{horse\})
(``skateboard'', \{skateboard\})
(``laptop'', \{laptop\})
(``cake'', \{cake\})
(``giraffe'', \{giraffe\})
(``table'', \{dining table\})
(``bench'', \{bench\})
(``motorcycle'', \{motorcycle\})
(``bathroom'', \{sink\})
(``elephant'', \{elephant\})
(``umbrella'', \{umbrella\})
(``kitchen'', \{oven\})
(``kite'', \{kite\})
(``people'', \{person\})
(``ball'', \{sports ball\})
(``sheep'', \{sheep\})
(``zebra'', \{zebra\})
(``phone'', \{cell phone\})
(``surfboard'', \{surfboard\})
(``hydrant'', \{fire hydrant\})
(``zebras'', \{zebra\})
(``teddy'', \{teddy bear\})
(``truck'', \{truck\})
(``stop sign'', \{stop sign\})
(``sandwich'', \{sandwich\})
(``boat'', \{boat\})
(``street'', \{car\})
(``bat'', \{baseball bat\})
(``bananas'', \{banana\})
(``giraffes'', \{giraffe\})
(``living'', \{couch\})
(``snow'', \{skis\})
(``bird'', \{bird\})
(``elephants'', \{elephant\})
(``vase'', \{vase\})
(``cows'', \{cow\})
(``broccoli'', \{broccoli\})
(``computer'', \{keyboard\})
(``woman'', \{person\})
(``tie'', \{tie\})
(``horses'', \{horse\})
(``bear'', \{bear\})
(``desk'', \{mouse\})
(``plane'', \{airplane\})
(``luggage'', \{suitcase\})
(``airplane'', \{airplane\})
(``person'', \{person\})
(``hot'', \{hot dog\})
(``refrigerator'', \{refrigerator\})
(``wii'', \{remote\})
(``kites'', \{kite\})
(``boats'', \{boat\})
(``couch'', \{couch\})
(``traffic'', \{traffic light\})
(``plate'', \{fork\})
(``surf'', \{surfboard\})
(``umbrellas'', \{umbrella\})
(``wine'', \{wine glass\})
(``skate'', \{skateboard\})
(``bowl'', \{bowl\})
(``stuffed'', \{teddy bear\})
(``room'', \{tv\})
(``cow'', \{cow\})
(``scissors'', \{scissors\})
(``snowboard'', \{snowboard\})
(``chair'', \{chair\})
(``car'', \{car\})
(``banana'', \{banana\})
(``bicycle'', \{bicycle\})
(``birds'', \{bird\})
(``vegetables'', \{broccoli\})
(``microwave'', \{microwave\})
(``donuts'', \{donut\})
(``video'', \{remote\})
(``batter'', \{baseball bat, person\})
(``skateboarder'', \{person, skateboard\})
(``surfer'', \{person, surfboard\})
(``skis'', \{skis\})
(``motorcycles'', \{motorcycle\})
(``meter'', \{parking meter\})
(``suitcase'', \{suitcase\})
(``sink'', \{sink\})
(``bike'', \{bicycle\})
(``chairs'', \{chair\})
(``food'', \{bowl\})
(``dogs'', \{dog\})
(``oven'', \{oven\})
(``court'', \{sports ball\})

%% file: assets/mt.tex
(``and'', \{und\})
(``Commission'', \{Kommission\})
(``not'', \{nicht\})
(``Union'', \{Union\})
(``we'', \{wir\})
(``I'', \{ich\})
(``that'', \{daß\})
(``Mr'', \{Herr\})
(``I'', \{Ich\})
(``Parliament'', \{Parlament\})
(``President'', \{Präsident\})
(``Member States'', \{Mitgliedstaaten\})
(``report'', \{Bericht\})
(``European'', \{Europäischen\})
(``We'', \{Wir\})
(``or'', \{oder\})
(``in'', \{in\})
(``Europe'', \{Europa\})
(``the'', \{der\})
(``Council'', \{Rat\})
(``between'', \{zwischen\})
(``is'', \{ist\})
(``2000'', \{2000\})
(``Commissioner'', \{Kommissar\})
(``EU'', \{EU\})
(``for'', \{für\})
(``the'', \{die\})
(``The'', \{Die\})
(``also'', \{auch\})
(``with'', \{mit\})
(``like to'', \{möchte\})
(``you'', \{Sie\})
(``1999'', \{1999\})
(``directive'', \{Richtlinie\})
(``only'', \{nur\})
(``proposal'', \{Vorschlag\})
(``European'', \{Europäische\})
(``Madam'', \{Präsidentin\})
(``Mrs'', \{Frau\})
(``Kosovo'', \{Kosovo\})
(``but'', \{aber\})
(``new'', \{neuen\})
(``Group'', \{Fraktion\})
(``have'', \{haben\})
(``behalf'', \{Namen\})
(``Mr'', \{Herrn\})
(``women'', \{Frauen\})
(``has'', \{hat\})
(``regions'', \{Regionen\})
(``years'', \{Jahren\})
(``all'', \{alle\})
(``two'', \{zwei\})
(``cooperation'', \{Zusammenarbeit\})
(``if'', \{wenn\})
(``1'', \{1\})
(``new'', \{neue\})
(``Article'', \{Artikel\})
(``because'', \{weil\})
(``whether'', \{ob\})
(``Parliament'', \{Parlaments\})
(``a'', \{eine\})
(``measures'', \{Maßnahmen\})
(``but'', \{sondern\})
(``institutions'', \{Institutionen\})
(``social'', \{sozialen\})
(``to'', \{zu\})
(``political'', \{politischen\})
(``development'', \{Entwicklung\})
(``national'', \{nationalen\})
(``today'', \{heute\})
(``countries'', \{Länder\})
(``European'', \{europäischen\})
(``must'', \{muß\})
(``our'', \{unsere\})
(``as'', \{wie\})
(``problems'', \{Probleme\})
(``initiative'', \{Initiative\})
(``work'', \{Arbeit\})
(``be'', \{werden\})
(``very'', \{sehr\})
(``human rights'', \{Menschenrechte\})
(``of the'', \{des\})
(``us'', \{uns\})
(``three'', \{drei\})
(``debate'', \{Aussprache\})
(``other'', \{anderen\})
(``hope'', \{hoffe\})
(``already'', \{bereits\})
(``question'', \{Frage\})
(``this'', \{diesem\})
(``debate'', \{Debatte\})
(``are'', \{sind\})
(``will'', \{wird\})
(``proposals'', \{Vorschläge\})
(``If'', \{Wenn\})
(``Prodi'', \{Prodi\})
(``Council'', \{Rates\})
(``rapporteur'', \{Berichterstatter\})
(``INTERREG'', \{INTERREG\})
(``role'', \{Rolle\})

%% file: assets/ec-vector-av.tex
(``15'', \{3\_3\})
(``25 25'', \{3\_0\})
(``3'', \{2\_3\})
(``20 20'', \{0\_3, 1\_0\})
(``7 7'', \{0\_3, 1\_3\})
(``4'', \{2\_0\})
(``16 16 16 16 16 16 16 16 16'', \{0\_3, 3\_0\})
(``2'', \{0\_0, 2\_0\})
(``13 13 13 13 13 13'', \{2\_0\})
(``23 23 23 23 23 23 23'', \{0\_0, 2\_3\})
(``28'', \{0\_0, 1\_3\})
(``27 27'', \{1\_0\})
(``17 17 17'', \{0\_0\})
(``31'', \{2\_0, 3\_2\})
(``22 22 22 22 22'', \{1\_3\})
(``22 25 25 25 25'', \{0\_1, 1\_3\})
(``30 30'', \{2\_1\})
(``22 22 13'', \{1\_3, 3\_3\})
(``26 26 26 26 26 26 26 26'', \{1\_3, 2\_0, 3\_0\})
(``15'', \{3\_1\})
(``15 27 27 27 27 27 27 27'', \{0\_1, 3\_2\})
(``8'', \{0\_0\})
(``3 3 3 30 30'', \{0\_1, 1\_1, 2\_2\})
(``16 16'', \{3\_0\})
(``3 3 3 3 30'', \{0\_2, 1\_2, 2\_2\})
(``7 7'', \{0\_2, 3\_1\})
(``15 3 3 3 3'', \{0\_2, 3\_2\})
(``15 7 20 27 27 27 27 27 27 27'', \{0\_2, 1\_1, 2\_1, 3\_2\})
(``20 27 27 27 27 27 27'', \{0\_2, 3\_2\})
(``15 15 15 3 27 27 27 27 27 27'', \{0\_2, 1\_1, 2\_2, 3\_2\})
(``22 22 22 22 22 30 30 30 30 30'', \{0\_1, 1\_2, 2\_2, 3\_2\})
(``8 1 23'', \{1\_1, 3\_0\})
(``22 22 22 25 3 30 30 30 30 30'', \{0\_1, 1\_2, 2\_2, 3\_1\})
(``28 28 2 2 2 2 2 2'', \{1\_2, 3\_1\})
(``22 22 22 17 17 17 17'', \{1\_2, 3\_3\})
(``26 26 6 4 4 4'', \{0\_1, 3\_0\})
(``23'', \{1\_0, 2\_3\})
(``15 15 15 15 15 15 15 15 15'', \{1\_2, 2\_3\})
(``22 22 22 3'', \{0\_1, 1\_2\})
(``7 7 20 20 20 20 20 20 20 20'', \{1\_1\})
(``7 7 7 7 7 20 7'', \{1\_2, 3\_2\})
(``31 31'', \{0\_3, 3\_3\})
(``28 28 28 8 8 8 12 12 12 12'', \{1\_2, 2\_1, 3\_0\})
(``15 15 15 15 15 17 17 17 17'', \{0\_1, 1\_2, 2\_2\})
(``3 27 27 27'', \{2\_2\})
(``7 15 15 15 15 15 15 15 15'', \{0\_3, 1\_2, 2\_2\})
(``3 3 3 3 3 3 3 3 3 3'', \{0\_2, 1\_0, 3\_0\})
(``7 13'', \{0\_1, 1\_2, 3\_2\})
(``28 26'', \{3\_0\})
(``15 15 7'', \{0\_2, 1\_3, 2\_2\})
(``22 22 23 23 23 23 23 23 23 23'', \{1\_1, 2\_2, 3\_1\})
(``7'', \{1\_2\})
(``13 13 13'', \{3\_3\})
(``7 7 7 7'', \{3\_2\})
(``15 17 17 17 17 17'', \{1\_1\})
(``22 22 22 17'', \{1\_2\})
(``15 15 15 13 13 13 13 13 17'', \{0\_1, 1\_2, 2\_1\})
(``3 3 3 3 3 23 23 23'', \{0\_1, 1\_1, 3\_1\})
(``7 16 16 16 16 4'', \{0\_3, 1\_1, 3\_1\})
(``26 26 26 26 6'', \{1\_2, 3\_0\})
(``22 22 22 22'', \{2\_2, 3\_2\})
(``25 25 25 25 25'', \{2\_2\})
(``7 25 25 25 25'', \{0\_2, 1\_3, 2\_3\})
(``22 7 26 13'', \{0\_1, 1\_3, 3\_2\})
(``15 15 15 15 17'', \{0\_1, 1\_2\})
(``23 23 17'', \{2\_2, 3\_2\})
(``7 26 26 26 26 26'', \{0\_2, 1\_3, 2\_1\})
(``8 8 8 23 23 23'', \{1\_2, 2\_2, 3\_1\})
(``7 26 26 26 26 26 26 26'', \{2\_0, 3\_1\})
(``22 7 26 26 26 26 28 28 28 28'', \{0\_1, 2\_1, 3\_1\})
(``15 15 15 15 15 15 15 15'', \{0\_2, 2\_3\})
(``5 4 4 4 4'', \{0\_2, 1\_0, 3\_1\})
(``26 26 26'', \{2\_0, 3\_0\})
(``22 13 13 13 13 13 13 13 2'', \{1\_2, 3\_2\})
(``15 15 31 31 31 31 31 31 31'', \{0\_2, 1\_1, 2\_1\})
(``22 28 28 28 28 28 28 28 28'', \{2\_1, 3\_1\})
(``15'', \{1\_1\})
(``13 13 13 13 2 2 2 2'', \{1\_1, 3\_2\})
(``1 1 1 1 1 1'', \{1\_2, 2\_3\})
(``8 8 30 30'', \{1\_1, 3\_0\})
(``4 4 4 27 27'', \{0\_1, 3\_1\})
(``17 17 17 17'', \{1\_0, 3\_3\})
(``23 23 23 23 23 23 23 27'', \{2\_2\})
(``15 31 31'', \{0\_2, 2\_1\})
(``5 27 27 27 27 27'', \{0\_2, 2\_1\})
(``22'', \{0\_0, 2\_3\})
(``28 8 8 8 8'', \{2\_2, 3\_0\})
(``17 2 2 2 2 2'', \{2\_1, 3\_2\})
(``22 22 2'', \{1\_1, 3\_1\})
(``3 3 23 23 23'', \{0\_1, 3\_1\})
(``28 28 28 28'', \{2\_1\})
(``26 26 26 4 4 4 4 4'', \{0\_1, 3\_1\})
(``17 17 27 27 27'', \{2\_1, 3\_2\})
(``15 13 13 13 13'', \{1\_2, 2\_1\})
(``25 3 25 3 25'', \{0\_1, 1\_2\})
(``20 20 20 27 27 27 27 27 27 27'', \{2\_1, 3\_1\})
(``3 3 3 3'', \{0\_2\})
(``31 31 31 31 31 31 31 31 31 31'', \{1\_0\})
(``17 13'', \{0\_0, 2\_1\})
(``3 3 3 3 3'', \{1\_0, 3\_0\})

%% file: assets/ec-vector-sparse.tex
(``22'', \{4, 7\})
(``24'', \{0, 3, 6\})
(``16'', \{1, 5\})
(``11'', \{6\})
(``26'', \{0, 4, 6\})
(``1 1'', \{3\})
(``6'', \{0, 7\})
(``17'', \{5, 7\})
(``28 28'', \{0, 2\})
(``18'', \{0, 2\})
(``21 21'', \{0, 6\})
(``14 14 14 14 14'', \{1, 3, 6, 7\})
(``16'', \{4, 7\})
(``24'', \{1, 5\})
(``1 28'', \{3, 4\})
(``31 31'', \{4, 7\})
(``12'', \{4\})
(``3'', \{4, 5\})
(``22 22 22'', \{0, 3, 6\})
(``4'', \{3\})
(``28'', \{2\})
(``12 12 12 12 12'', \{2\})
(``28 27 27 27 27 27 27'', \{0, 4\})
(``28 9'', \{0, 3, 4\})
(``11 11 11'', \{3\})
(``7'', \{3\})
(``12 12 12 12'', \{2\})
(``24 5'', \{4\})
(``30 30'', \{1, 7\})
(``14 14 14 14'', \{1, 2, 7\})
(``25 25 25 25 25 25'', \{1\})
(``4 5 5 5 5'', \{2, 4\})
(``26'', \{2\})
(``25 25 27 27 27 27 27'', \{0, 5\})
(``1 29 29 29 29 29 29 29 12 12'', \{1, 3, 6\})
(``5 5 5 5 5 5 5 5'', \{4\})
(``1 1 1 1 1 1 1'', \{1, 7\})
(``1 12 12 12 12 12 12 12 12'', \{1, 3\})
(``6 6 6 6'', \{2\})
(``1 1 1 1'', \{0, 1\})
(``1 30'', \{2\})
(``1 18'', \{3, 5\})
(``23'', \{7\})
(``16 16 16 16'', \{0\})
(``21 21 21 21'', \{7\})
(``5 5 5 27 27'', \{1\})
(``4 5'', \{4\})
(``5 12'', \{2, 3\})
(``1 12'', \{3, 5\})
(``18 27'', \{5\})
(``11 22 22'', \{2, 3\})
(``22'', \{6\})
(``22 22 11'', \{1, 2\})
(``4 4 27 27'', \{5\})
(``1 29 29'', \{3\})
(``16 16'', \{0\})
(``12'', \{1, 3\})
(``9 9 9 9 9 9 9 9'', \{0\})
(``6 1 1 1'', \{1, 2\})
(``20 20 20 20 20 20 20 20'', \{1, 7\})
(``26 26 26 26 28 28 28'', \{1, 7\})
(``29 29 4 12 12 12 12 12 12'', \{5\})
(``1 1'', \{1, 2\})
(``29 29 4 4 4 4 4 4 4 4'', \{1, 6\})
(``11 11 23 23 23'', \{3\})
(``12 12 12'', \{2\})
(``22 22'', \{3\})
(``28 12'', \{0\})
(``21 21 21 8'', \{2\})
(``1 4 4'', \{2\})
(``21 21 23'', \{2\})
(``21 21 21 21 21 21 21 21 6'', \{1, 2\})
(``12 16'', \{2\})
(``10 10 10 10 10 10 10 25 25 25'', \{2, 7\})
(``28 28 28'', \{1, 4\})
(``24 24 24 24 24 24 24 24'', \{7\})
(``21 21 6 6'', \{2\})
(``21 21'', \{2\})
(``9 5'', \{4\})
(``31 31 31 31'', \{3\})
(``28 28 28 28 28 28 28'', \{3\})
(``10 10 10 10 10'', \{7\})
(``13 13 13'', \{7\})
(``11 14 14 22 22 22 17 17 17'', \{2\})
(``7 7 7 7'', \{2\})
(``22 26 26 5 5 5 5 5 5'', \{1\})
(``22 22 22'', \{2, 5\})
(``11 11'', \{7\})
(``12 12 27 27 27 27 27'', \{2\})
(``30 30 30 30 30 30 27 27 27 27'', \{5\})
(``2 9 12 12'', \{2\})
(``11 11 23 17 17 17'', \{3\})
(``18 18 18 18 18 18 28 18'', \{1\})
(``25'', \{0\})
(``23 23 23 23 19 19 19 19'', \{1\})
(``6 6 10 10 10 10 10 10 10'', \{3\})
(``6 6 6 6 6 6 6 17 17'', \{3\})
(``24 24 24 4 4 12 12'', \{2\})
(``22 22 22 22 28 28 5'', \{1\})
(``24 4 4 4 4 4 4 4 4 4'', \{2\})

%% file: assets/ec-shapeworld-ref.tex
(``29'', \{ellipse\})
(``29'', \{gray\})
(``29'', \{green\})
(``29'', \{rectangle\})
(``29'', \{triangle\})
(``29'', \{white\})
(``30 2'', \{ellipse\})
(``30'', \{square\})
(``29 29'', \{blue\})
(``18 4 18'', \{white\})
(``29 29'', \{circle\})
(``5 3'', \{square\})
(``5 3'', \{rectangle\})
(``6'', \{square\})
(``24 18'', \{ellipse\})
(``11 4'', \{white\})
(``30 5'', \{triangle\})
(``2 2 2 2 2'', \{white\})
(``29'', \{yellow\})
(``11'', \{ellipse\})
(``2 2 3'', \{ellipse\})
(``4'', \{rectangle\})
(``30 30 3'', \{ellipse\})
(``2'', \{square, yellow\})
(``30'', \{circle\})
(``2 2 2 2 2'', \{ellipse\})
(``3'', \{gray\})
(``23 5'', \{rectangle\})
(``4'', \{green\})
(``13 6 13 2'', \{ellipse\})
(``23 18 23 23'', \{white\})
(``3'', \{rectangle\})
(``18 3 18'', \{white\})
(``2 2 5'', \{ellipse\})
(``24 6 24 6'', \{white\})
(``6 2'', \{ellipse\})
(``3'', \{green\})
(``13 13 23'', \{square\})
(``24 24'', \{white\})
(``18 18 18 23 18'', \{white\})
(``30 5 2'', \{ellipse\})
(``23 24 23'', \{ellipse\})
(``18 4 18 5'', \{ellipse\})
(``4 18 4 5'', \{yellow\})
(``13'', \{gray\})
(``18 5 4 18'', \{ellipse\})
(``2 18'', \{white\})
(``4 5 4'', \{ellipse\})
(``18 18'', \{yellow\})
(``23 13 23'', \{square\})
(``6 3'', \{triangle\})
(``23 3 23 3 23'', \{yellow\})
(``13 6 13 24'', \{ellipse\})
(``24 24'', \{yellow\})
(``13 13 24 13 13 13 24'', \{ellipse\})
(``6 3'', \{circle\})
(``23 18 23'', \{ellipse\})
(``2 2 23 2'', \{white\})
(``5 23'', \{green\})
(``30 30'', \{red\})
(``18 5 4'', \{yellow\})
(``3 23 3 3 3'', \{square, yellow\})
(``23 24 23'', \{white\})
(``18 18 3 18'', \{ellipse\})
(``3 3 3 3'', \{square\})
(``24 6 13 6'', \{white\})
(``30 6 30 3'', \{ellipse\})
(``18 3 18'', \{yellow\})
(``5 30'', \{blue\})
(``24 2'', \{ellipse\})
(``24 13 24 13 13'', \{square, white\})
(``23 18 23'', \{square, yellow\})
(``4 18 18 4'', \{ellipse\})
(``2 2 3'', \{white\})
(``13 6 13'', \{ellipse\})
(``13 13 24'', \{white\})
(``18 23 18'', \{white\})
(``13'', \{rectangle\})
(``13 13 24 13 13'', \{ellipse\})
(``30 24 30'', \{white\})
(``13 13 13 13'', \{white\})
(``23 3 23 3 23'', \{ellipse\})
(``5 18 18 5'', \{ellipse\})
(``24'', \{green\})
(``13 13 23 13 13 13 24'', \{circle, red\})
(``30 6 30 6'', \{blue\})
(``5 5 4'', \{ellipse\})
(``13 24'', \{blue\})
(``5'', \{circle, red\})
(``30 6 30 2'', \{white\})
(``3 3 3 3 3 3'', \{yellow\})
(``18 5 18 5 18'', \{ellipse\})
(``3 3 3'', \{white\})
(``18'', \{square\})
(``18 5 18 4 2'', \{ellipse\})
(``13 2'', \{ellipse\})
(``3 3 23 3 3 23 3 3 2'', \{circle, red\})
(``18 3 18 5 18 3'', \{ellipse\})
(``4 4'', \{blue\})
(``13 24 13'', \{yellow\})

%% file: assets/ec-shapeworld-setref.tex
(``3 3'', \{circle, not\})
(``21 21'', \{circle\})
(``23 23'', \{gray, not\})
(``20 20'', \{blue, not\})
(``5 5'', \{and, green, not\})
(``28'', \{square\})
(``26'', \{or, yellow\})
(``28'', \{ellipse, not\})
(``4 4'', \{white\})
(``11'', \{not, rectangle\})
(``11 11'', \{ellipse\})
(``25 25 25'', \{blue, red\})
(``25 4'', \{blue\})
(``3 28'', \{triangle\})
(``22 26'', \{not, red\})
(``25 23'', \{green, or, red\})
(``5 4 5'', \{gray, or, white\})
(``12 23'', \{yellow\})
(``12 18'', \{or, red, white\})
(``23 25'', \{and, gray, white\})
(``5 20'', \{gray, or, red\})
(``4 23'', \{and, red\})
(``12 12'', \{yellow\})
(``3 11'', \{and, circle\})
(``21'', \{circle, or\})
(``28'', \{rectangle\})
(``23 26 23'', \{green\})
(``26 20'', \{and, white\})
(``11'', \{ellipse\})
(``21 22'', \{and, triangle\})
(``22 5'', \{blue, green\})
(``28 25'', \{triangle\})
(``5 26 5'', \{gray\})
(``3'', \{and, circle\})
(``25 20'', \{white\})
(``25 26 4'', \{blue\})
(``28 3'', \{triangle\})
(``21'', \{square\})
(``12'', \{yellow\})
(``11 22'', \{triangle\})
(``12 25'', \{or, red, white\})
(``21'', \{and\})
(``3'', \{square\})
(``20 12 20 12'', \{blue, not\})
(``20 22 22'', \{or, triangle, white\})
(``18'', \{rectangle\})
(``5 5'', \{gray\})
(``5'', \{red\})
(``23 22'', \{red\})
(``23 23'', \{green\})
(``26 22'', \{and, white\})
(``3'', \{rectangle\})
(``20 5 5'', \{white\})
(``22'', \{or\})
(``5 5'', \{triangle\})
(``12 12 23'', \{not\})
(``27'', \{triangle\})
(``12 5'', \{green, not\})
(``25 23 20'', \{and, gray, white\})
(``25 21 4'', \{blue\})
(``4 25 12'', \{blue, or\})
(``22 22'', \{triangle\})
(``20 3 20'', \{white\})
(``22 20'', \{blue, not\})
(``12 22'', \{triangle\})
(``28 26'', \{triangle\})
(``21 23'', \{green\})
(``20 20 20'', \{and, white\})
(``25'', \{blue\})
(``28 20'', \{triangle\})
(``22 22'', \{not\})
(``5'', \{gray\})
(``12'', \{or\})
(``5 11'', \{and, green\})
(``3'', \{triangle\})
(``4 22'', \{red\})
(``23'', \{not\})
(``23 4'', \{green\})
(``18'', \{triangle\})
(``27'', \{rectangle\})
(``12 20 12 20 23'', \{and\})
(``20 22'', \{blue\})
(``25 23'', \{and, gray\})
(``4 12 23'', \{and\})
(``23 3 23'', \{green\})
(``20 22 20'', \{white\})
(``12 20 23'', \{and\})
(``12 20 12 20'', \{and\})
(``8 5 12'', \{and, not\})
(``23 11 23'', \{green\})
(``23 20'', \{white\})
(``28 4 28 4 25'', \{and, triangle, white\})
(``21'', \{triangle\})
(``25 5 20'', \{white\})
(``22 26'', \{gray\})
(``28 4'', \{triangle\})
(``26 18'', \{and\})
(``5 4 18'', \{white\})
(``12 20 12'', \{and\})
(``28'', \{or\})

%% file: assets/ec-shapeworld-concept.tex
(``3 6'', \{gray, not\})
(``7 7'', \{blue, not\})
(``32'', \{circle\})
(``4 5'', \{not, yellow\})
(``6 12 6 12'', \{green, or, yellow\})
(``3 12 3'', \{blue, or, yellow\})
(``3 7 3 3'', \{green, white\})
(``6 6'', \{not, red\})
(``4 7 4'', \{green, or, red\})
(``6 28 6 28'', \{or, white, yellow\})
(``25 5 25'', \{blue, or, white\})
(``32 32 32'', \{ellipse\})
(``5 5 5 5 5 5 5 5'', \{green, not, yellow\})
(``25 25 25 25 25'', \{green, not, red\})
(``3 4 3'', \{and, white, yellow\})
(``28 28 28 28'', \{white\})
(``32'', \{rectangle\})
(``5 3 5'', \{blue, red\})
(``12 28'', \{yellow\})
(``22'', \{square\})
(``22'', \{triangle\})
(``7 28'', \{or, red, yellow\})
(``3 6 3'', \{white\})
(``5 32'', \{or, red, white\})
(``28 28 5'', \{gray, or, white\})
(``7 28 7 28 7 28 7'', \{blue, green\})
(``3 31 3'', \{blue\})
(``12 4 12'', \{and, green\})
(``28 3'', \{gray\})
(``5 6 5 6 5'', \{and, red, yellow\})
(``25 7 25'', \{green, not, or\})
(``7 5 7 5 7 5 7 5'', \{blue, not, yellow\})
(``4 4 4'', \{not, or, yellow\})
(``22'', \{and, ellipse, not\})
(``6 12 6 6'', \{and, gray\})
(``31 31 31 31'', \{and, blue\})
(``7 12 7'', \{and, white\})
(``5 5 3 28 7'', \{gray, or, red\})
(``28 28 31'', \{gray\})
(``25 3'', \{red, white\})
(``7 6 7 6 7'', \{blue, not, or\})
(``32'', \{triangle\})
(``22 22'', \{or, rectangle\})
(``12 12 32'', \{and, yellow\})
(``5 28'', \{red\})
(``4 6 4'', \{or\})
(``5 7 5 7'', \{and, yellow\})
(``32'', \{and, square\})
(``4 12'', \{green\})
(``3 3 3 3'', \{and, not\})
(``3 3 6'', \{and\})
(``32 32'', \{ellipse, or\})
(``6 5 5'', \{and, gray\})
(``7'', \{or, red\})
(``4 3'', \{and, gray\})
(``12 12 12'', \{yellow\})
(``28 7 28 7'', \{and, green\})
(``25 25 25'', \{and\})
(``28 3 3'', \{or\})
(``23 23'', \{blue, not\})
(``4 4 32 4 32 4'', \{not, or, yellow\})
(``4 3 4 3'', \{not\})
(``6 4'', \{or\})
(``25 7 32'', \{green, not\})
(``31'', \{blue, or\})
(``4'', \{not\})
(``5 5 5 25'', \{green, not, yellow\})
(``28 31 28 31'', \{gray\})
(``12 22 12'', \{rectangle, yellow\})
(``3 3 3'', \{and, not\})
(``5 3 5'', \{or\})
(``28'', \{or, white\})
(``5 5 3'', \{gray\})
(``31 28'', \{red\})
(``6 25'', \{not, red\})
(``32 27 27'', \{ellipse, yellow\})
(``7 6 7 32 7'', \{blue, not\})
(``32 6'', \{not\})
(``4'', \{green, or\})
(``12 6 28'', \{yellow\})
(``32'', \{ellipse\})
(``27 27'', \{yellow\})
(``25 32'', \{not\})
(``3 6 7'', \{white\})
(``28 28 12 3'', \{yellow\})
(``4'', \{circle\})
(``27 32 27'', \{yellow\})
(``3 4'', \{and, gray\})
(``6 7 6'', \{square\})
(``6 12 25'', \{red\})
(``23 22 23'', \{blue\})
(``6 3'', \{white\})
(``3 3 7'', \{green, white\})
(``3 5 5'', \{or\})
(``5 27 32'', \{or, red, white\})
(``7'', \{blue, not\})
(``5'', \{and, yellow\})
(``7 12'', \{and, white\})
(``28'', \{gray\})
(``3 25 3'', \{not\})